# Machine Learning Algorithms for Time Series Analysis and Forecasting

Rameshwar Garg
*CSE Department*
RV College of Engineering
Bangalore, India
ramgarg52@gmail.com

Shriya Barpanda
*ECE Department*
RV College of Engineering
Bangalore, India
shriyabarpanda@gmail.com

Girish Rao Salanke N S
*CSE Department*
RV College of Engineering
Bangalore, India
girishraosalanke@rvce.edu.in

Ramya S
*ECE Department*
RV College of Engineering
Bangalore, India
ramyas@rvce.edu.in

*Abstract*—Time series data is being used everywhere, from sales records to patients' health evolution metrics. The ability to deal with this data has become a necessity, and time series analysis and forecasting are used for the same. Every Machine Learning enthusiast would consider these as very important tools, as they deepen the understanding of the characteristics of data. Forecasting is used to predict the value of a variable in the future, based on its past occurrences. A detailed survey of the various methods that are used for forecasting has been presented in this paper. The complete process of forecasting, from preprocessing to validation has also been explained thoroughly. Various statistical and deep learning models have been considered, notably, ARIMA, Prophet and LSTMs. Hybrid versions of Machine Learning models have also been explored and elucidated. Our work can be used by anyone to develop a good understanding of the forecasting process, and to identify various state of the art models which are being used today.

## I. INTRODUCTION

In today's world, data is always being recorded and can be used for a wide variety of purposes. Many important decisions can be made with this data as the foundation. Data generated from website traffic, the stock market, daily seasonal effects, sales reports, business reports and many other sources can be categorised into time series data. Time series data is an ordered collection of data samples, each denoting an event occurring at a specific time. The analysis and forecasting of time series data forms an integral part of Data Science and Machine Learning (ML) and has proven to be extremely useful in providing crucial insights while making business decisions. Many characteristics of data that go unnoticed, and other meaningful statistics, can be derived with the help of Time Series Analysis (TSA). TSA is a precursor of Time Series Forecasting (TSF). Forecasting is the use of historical data, to gain insights and make predictions about the future. This task of forecasting is not a trivial one, as data from the future is always unavailable, and can be estimated only from events that have already occurred. TSF is an intricate process that involves the study and handling of many problems that can be associated with the data. If carried out effectively, TSF can be very profitable, both in terms of information and revenue.

As is the case with most ML techniques, the quality of TSF improves with the quality and quantity of data. A large amount of data provides more opportunities for Exploratory Data Analysis (EDA), which in turn leads to the creation of more robust models which can be trained, tested and tuned effectively [1]. The accuracy of the forecast also depends on the period or horizon for which the forecast is being made. The forecasts can be classified into short term, medium term and long term forecasts. Generally, short term forecasts can be made more easily and with lesser uncertainty. Forecasts can also be improved with timely updates. If forecasts are repeated with more recent data, they tend to be more accurate than before. Another factor which affects the forecast, is the frequency at which we make the forecast, i.e., whether the forecasted data is timely with an interval of a year, a month, a day, an hour and so on.

TSF is a combination of multiple processes, all of which involve the usage of time series data in different ways. The first broad process is Data Preparation. In this phase, the data is cleaned, preprocessed and formatted in such a way that it can be used for forecasting. Feature Engineering can also be applied on the data, to make the more data more suitable for the learning model and increase the opportunity for the model to learn. The next phase involves the creation of the model followed by the application of the model on the preprocessed data in order to generate the forecasts. The final and arguably the most important phase is the validation of the model and the forecasts. Validation is one of the most important steps in any ML process, and more so in TSF as the forecasted data needs to be evaluated, in order to be considered useful.

In this paper, we present a survey of various TSF methodologies, with a focus on the second phase of the entire process. The next section contains information regarding the cleaning and preprocessing of time series data. The third section focuses on the explanation of various statistical and deep learning models that can be used for forecasting , while presenting a thorough survey on the same. In section four, multiple error metrics have been touched upon, with a detailed explanation on the validation of the model. Finally, we end our paper with section five, where we present the conclusion.

## II. DATA PREPARATION

Data preparation involves a lot of components and factors when time series data is considered. Depending on the nature of the data, various techniques have to be applied in order to make the data suitable for forecasting. In the following subsections, we have described the various processes that are involved in the data cleaning and preprocessing of time series data.

### A. Missing Value Imputation

Missing value imputation is the technique to handle missing values in the data. The missing values can be replaced by other meaningful values by a variety of methods. A mean value of the entire data can be calculated, and then used to replace all the missing values. Other methods to replace the missing values are to carry forward the previous value, or bring back the next value. These two methods are generally considered as the newly added data would then fit into the pattern being followed at that point of time.

### B. Data Resampling

The next step in the data preparation stage involves upsampling or down-sampling the data. Up-sampling is the process of increasing the frequency of the data samples, i.e., converting daily data into hourly data and so on. When the data is upsampled, missing values are generated and imputation needs to be carried out again. On the other hand, downsampling is the process of reducing the frequency of the data samples, by taking a mean of, or finding the maximum value from the unsampled data. Data can be converted from hourly to daily or even monthly data in the case of down-sampling. The data should be sampled keeping in mind the use case and the requirements.

### C. Data Slicing

Selecting the appropriate amount of data that needs to be considered, from a starting time to an ending time is known as data slicing. A data slice must be selected based on the requirements and even the characteristics of the data that the model is expected to learn. If the data before a particular time has become irrelevant, due to some reasons, that data can be excluded with the help of data slicing. Generally, slicing is carried out cautiously as any data is useful data.

### D. Data Visualisation & Decomposition

The most critical decisions regarding any time series data can be made only after the data is visualised properly. Analysing the data from a visual point of view, helps unearth many hidden patterns and characteristics of the data. Time series data comprises four main components, which can be identified visually and can be confirmed with the help of data decomposition [2]. These components are:

- Level: The level of time series data is the base value of the time series. When the data is averaged and expressed as a straight line, the level is obtained.
- Trend: The linear upward and downward movement of the data over a period of time is known as the trend of the time series data.
- Seasonality: The patterns or cycles which repeat with time, make up the seasonality of the data.
- Noise: The random variability in the data, which cannot be elucidated is known as noise.

These components, when grouped together, form the original/observed time series data. The data can either have additive trend and seasonality, or multiplicative trend and seasonality. An additive model is obtained by adding the trend, seasonality and residuals, while a multiplicative model is the product of the same. Examples of both can be seen in Figure 1. For an Additive model,

$$Y_t = T_t + S_t + R_t \qquad (1)$$

For a Multiplicative model,

$$Y_t = T_t * S_t * R_t \qquad (2)$$

Where, $Y_t$ is the resultant/observed time series data at time $t$, $T_t$ is the trend at time t, $S_t$ is the seasonality at time $t$ and $R_t$ denotes the residuals or errors at time $t$.

### E. Induction Of Stationarity

Time series data is said to be stationary when the mean, variance and covariance of the data are constant, and do not change as functions of time. Many forecasting models require the data to be stationary, as they assume the lack of trend and seasonality. Various tests can be employed to check whether the data is stationary. Plotting the Autocorrelation and Partial Autocorrelation plots is one of the best methods to identify non-stationarity. The Augmented Dickey-Fuller (ADF) test and the Kwiatkowski–Phillips–Schmidt–Shin (KPSS) test are two statistical tests to verify whether the data is stationary or not.

Once the lack of stationarity has been identified, the data can be transformed in order to induce stationarity. One of the most common methods to induce stationarity is to difference the data, by subtracting the value at the previous time step, from the current value. Seasonal differencing can also be performed, where the data in the previous season is subtracted from the current data, in order to remove seasonality, and induce stationarity. Various transformations such as logarithmic transformation, power transformation and smoothing can also be applied to induce stationarity. Once the data has been made stationary, the next phase of fitting a model and making forecasts can begin.

## III. MODEL FITTING AND FORECASTING

Various models can be employed to find patterns in the data, and then make predictions for the future, based on the history. These models can either have a statistical base or can be based on Deep Learning (DL) methodologies. In this section, we present a thorough review of various methodologies to forecast time series data.

### A. Statistical Models

Statistical models for TSF predict values based on classical methods, without the use of neural networks. These models tend to have lower accuracy, and are generally incapable of forecasting complex patterns. Some models such as Prophet perform better than the others, and can be employed for business use cases as well. These models should be used when the training data can change fast, and the time for a forecast becomes more important than the overall accuracy.

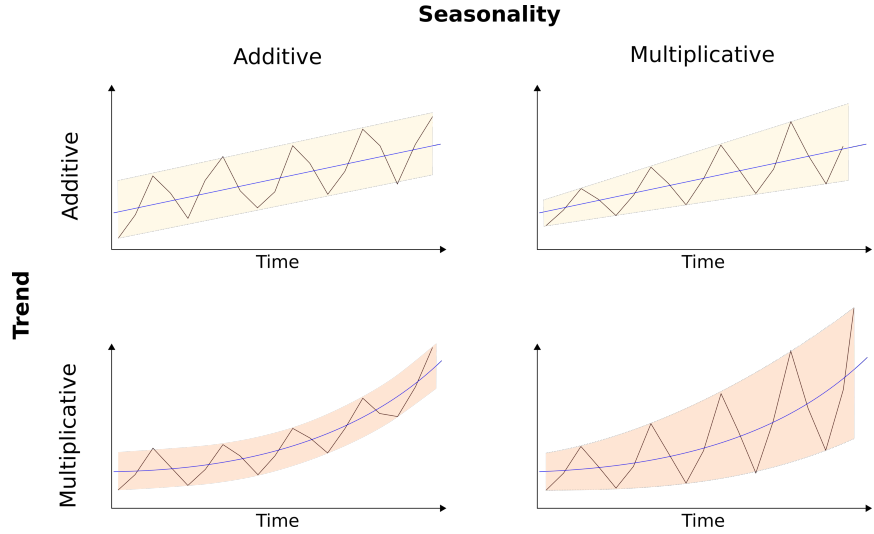

Fig. 1: Representation of Time Series Data with Different Types of Trend & Seasonality [3]

*1) Simple Exponential Smoothing (SES):* The SES model employs the logic of using weighted averages of historic data, to forecast the future. In this model, more importance is given to recent values, as they are assigned higher weights. The weights for older data samples keep decreasing with time. The model assumes the lack of any trend or seasonality, and can produce a linear forecast. The model requires a single parameter, $\alpha$, which is used as the relative weight. This parameter is known as the smoothing factor or the smoothing coefficient, and lies between 0 and 1.

$$\hat{y}_{T+1|T} = \alpha y_T + \alpha(1-\alpha)y_{T-1} + \alpha(1-\alpha)^2 y_{T-2} + \ldots \quad (3)$$

or,

$$S_{T+1} = \alpha y_n + (1-\alpha)S_t \quad (4)$$

In the above equations, $\hat{y}_{T+1|T}$ is the forecast at time $T+1$, $\alpha$ is the smoothing factor, $y_T$ is the value of the series at time $T$. Also, $S_{T+1}$ is the same as $y_{T+1|T}$. This model was presented by Charles C. Holt in 1960 [4], as an extension of the work of Robert Brown. This model can be used for forecasting successfully, and has been done so in a variety of works.

Eva and Oskar have presented their work, in which they use SES for forecasting [5]. In their paper, an end to end forecasting procedure has been explained and carried out, with thorough validation of their model. They have also explained how the forecasts vary with the change in values, and have conducted a detailed analysis of the same. They have also identified several drawbacks of SES and have explained them well. In [6], the authors have shown that the SES model can produce good seasonal forecasts as well, provided the model is tweaked properly. The use of SES results in good forecasts, with low error scores, and the model outperforms other contemporaries with higher complexity and prediction power. This shows that different types of data require different preprocessing and different models, and understanding the nature and characteristics of the data is an absolute must, before any model can be used for forecasting. Sulandari et al. have extended the standard model for exponential smoothing by building an Exponential Smoothing model which can make accurate forecasts, even when the data has multiple seasonalities [7]. They have also conducted a thorough survey and have visually analysed the results in depth. The authors seek to build a model that can fit multiple use cases and cater to a wide range of functionalities.

*2) Holt - Winters:* The Holt - Winters model, also known as triple exponential smoothing, is a forecasting method used when the data has both trend and seasonality. The capability of recognising and dealing with the seasonality component has been added as an upgrade, over the double exponential smoothing, or the Holt - Linear model, which assumes that the data has a trend. The seasonality can either be additive or multiplicative. This model makes use of three coefficients, for the level, for the trend and for the seasonality. The resulting forecast can be expressed as:

$$F_{T+k} = L_t + kT_t + S_{t+k-M} \quad (5)$$

where, $F_{t+k}$ is the forecast at time $t$ after taking $k$ steps into the future, $L_t$ is the Level estimate at time $t$, $T_t$ is the Trend estimate at time $t$, $M$ denotes the number of seasons and finally $S_t$ is the Seasonal estimate at time $t$. Here, $L_t$ makes use of $\alpha$, $T_t$ makes use of $\beta$ and $S_t$ makes use of $\gamma$. The Holt Winters model can be employed for a plethora of use cases, and generates impressive forecasts. It can also be coupled with various other techniques to increase the overall accuracy of the predictions being made. The initial model was developed by Charles Holt and Peter Winters and had been presented by C. Chatfield [8] in 1978. Various extensions of this model have been made and are being used today, to achieve desirable results.

Lifeng et al. have combined the grey accumulated generation technique with the Holt Winters model, to be able to handle the non stationarity of the irregular and random series, and even the seasonal effects that are visible in the data [9]. The study has been directed towards the forecast of the Air Quality Index, which has irregular data. They have made use of the fraction order accumulation methods to overcome this issue. Thorough tuning is required to achieve the optimal results, and this can be achieved after conducting a good amount of experiments. In [10], the authors have proposed a method to tweak the , and parameters automatically. Since data tends to be insufficient and irregular when real time data is considered, they have combined the fruit fly optimisation algorithm with the Holt - Winters model, to automatically select the smoothing parameters. The proposed approach has proven to be effective even when a small training set is available, and the model is computationally inexpensive. The authors of [11] have made an ensemble model using game theory models and the Holt Winters model. Their aim is to detect and prevent anomalies in Software Defined Networks. They achieve this goal using the anomaly detection capabilities of the Holt Winters model by making forecasts, and an autonomous decision making model based on game theory. Their results were satisfactory and proved that the Holt Winter model can be used contextually, i.e., to achieve different goals for different applications.

*3) ARIMA:* ARIMA is one of the most famous models for TSF, because of its high power and simplicity. It is used extensively and has many functionalities. There are many varieties of ARIMA, each for a slightly different use case. ARIMA is a forecasting model that considers the temporal nature of the data and fits a regression line to the data, where the future is directly dependent on the past. ARIMA when expanded, results in Auto Regressive Integrated Moving Average.

'AR', or Auto Regressive implies that the changing value, or the variable being predicted, regresses on its past/prior/lagged values. This means that the future can be obtained using a weighted combination of the past. The 'I' term, which stands for Integrated, is used to denote differencing that takes place within the model, to reduce the seasonality. 'MA' stands for Moving Average, and means that the value of the dependent variable, for the future, is a function of the residual error terms from the past. Furthermore, an integer set $(p,d,q)$ is used as the parameters set for the ARIMA model. Here, $p$ denoted the lag order, or the number of lags that are considered, $d$ represents the degree of differencing and finally $q$ is used to portray the number of moving average terms that have been considered.

The ARIMA model was proposed by George Box and Gwilym Jenkins in 1970 [12] and has been improved significantly since then. A novel method [13] to forecast the spread of COVID-19 has been proposed by Maher et al. using ARIMA. The proposed method is a hybrid dynamic model based on SEIRD, which stands for Susceptible Exposed Infected Recovered Dead, with automatically selected parameters for ascertainment rate. The residuals of the results are then corrected using ARIMA. Both long and short term forecasts can be obtained with confidence intervals. Thorough validation has been done and the hybrid model has proven to be effective. In [14] a hybrid model has been proposed, combining various other statistical ML models with ARIMA to generate forecasts. ARIMA is used to identify the linear relationships, and Support Vector Machines (SVM), Random Forest (RF) and Neural Networks to identify and forecast the nonlinear components. Various combinations of the linear and nonlinear models were used and tested adequately. The results were surprisingly good and the time complexity of the model is also acceptable. Thejas et al. have proposed the use of an AR and MA model to forecast the occurrence of click fraud, for multivariate time series [15]. A novel method has been used, where the tuning is succeeded by the minimisation of the Bayesian Information Criteria and the Akaike Information Criteria to find the model that performs the best for multi time scale data. The proposed algorithm has been validated using multiple error metrics and a significant amount of visual analysis has also been carried out.

*4) Prophet:* Prophet is a time series forecasting model that has been created by Facebook. The idea behind developing this model is to help any beginner in the field achieve some basic forecast, while also having the capability to forecast and predict complex time series and find intricate patterns. Facebook's Prophet, also known as fbProphet, uses a General Additive Model (GAM) for time series forecasting. In this type of model, the model is trained on the overall trend, and other effects such as seasonality and holidays are added to it. Prophet can accommodate a plethora of data characteristics in its model and can make accurate forecasts for any kind of data. Prophet also serves as an end to end forecasting model by itself, as it has the added functionality of cross validation where backtesting is performed to gauge the model, and several performance metrics are used to evaluate the model. Furthermore, it has the added functionality of hyperparameter tuning itself, in order to make sure that the best forecasts are being generated.

The fbProphet model fits on the data, assuming a decomposable time series, containing the trend, seasonality, holidays and errors. Its equation can be seen below.

$$f(t) = g(t) + s(t) + h(t) + e(t) \qquad (6)$$

In the above equation, *f(t)* represents the forecasting function, *g(t)* is used to model the trend function, which can be in the form of a linear or logistic function, *s(t)* represents the changes that occur seasonally, with a seasonal period of a week, a month, a year, or even all of them, *h(t)* is used to portray the and generate predictions. These models can take time to train, but are more adept at exploiting data to predict complex patterns which are not easily found. DL based models have higher accuracy and can be tuned to a greater extent as compared to the statistical models. Such models should be effects of holidays that occur randomly and have irregular effects and finally *e(t)* represents the error terms. The model has been proposed recently and is still gaining popularity. It has been proposed by Sean Taylor and Benjamin Letham [17] in 2018 and is already being used by many businesses. The implementation is available in Python and R and has been well documented.

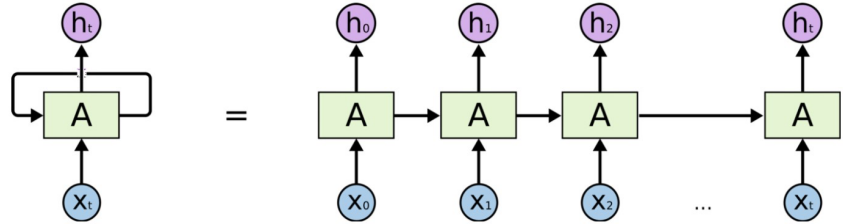

Fig. 2: An Unrolled Recurrent Neural Network [16]

In [18], the authors have proposed the usage of Prophet to forecast the prices of various cryptocurrencies. The fluctuation of these prices is very random and extremely difficult to predict. Since multiple cryptocurrencies have been used, the variety of the data is quite vast and fitting the model on this data is a non-trivial task. Although it seems improbable, the model has found the pattern really well and has forecasted the value of the currencies extremely accurately. This shows that Prophet can be used for TSF when the data varies very randomly as well. Battineni et al. have employed Prophet in yet another use case, where the pattern is quite unpredictable [19]. Based on their study, it is evident that Prophet is a perfect model for nonlinear trends that experience multiple types of seasonalities. Especially for the forecasting of the spread of a virus, and the extension of the pandemic, this model is suitable because additional parameters such as lockdowns and travel bans can also be incorporated as holiday effects, which significantly affect the forecast. A hybrid of Prophet has been proposed in [20], where the model combines Prophet and adaptive Kalman filtering. Initially, a modified version of the grey analysis method has been used to capture information regarding the relevant influencing features. This is followed by the use of the combination of the adaptive Kalman filter and fbProphet to predict various customers' maximum power demand. The results of forecasting had improved considerably and the method can be successfully used for this use case.

*B. Deep Learning Models*

Forecasting models which make use of DL methods, employ neural networks to identify and learn the underlying patterns employed when the accuracy of the forecast is of prime importance, and the time for training the model adequately can be spared.

*1) Convolutional Neural Networks (CNN):* For time series with multiple features, features have to be engineered and extracted from the time series data, which can be quite a challenging task, even for experts in the particular domain. CNNs can be used to learn features and make forecasts. For a time series of length *n* and width *k*, the kernel of the CNN will have a width of *k*, and the length can be varied as per the desired size of the window. As a result, the kernel can move only unidirectionally, from the beginning of the time series towards the end, while it performs convolution. Subsequent operations are applied using the kernel, including filtering and max pooling to achieve the final forecast.

Mehtab, Sen and Dasgupta have proposed various models using CNNs to forecast the prices of stocks in the stock market [21]. The forecasts obtained by these models were of a very high quality, but were computationally expensive. For this use case, with this data and model parameters, the CNN model outperformed the more complex DL models. The regression results were surprisingly pleasing, which proves that CNNs can be used effectively for time series forecasting. Xue et al. have proposed a combination of the CNN and the Long Short Term Memory (LSTM) model to achieve accurate forecasts [22]. In their proposed approach, a technique to automate the searching of the optimal model architecture has also been proposed. The various experiments that were conducted prove that the model can be used to forecast complex, nonlinear time series data quite efficiently. In [23], another method making use of the CNN-LSTM model has been proposed. This time the method has been used to forecast the price of gold, at a global level. Their experiments proved that combining CNNs with high complexity models such as LSTMs makes the hyperparameter tuning a very sensitive matter, which can lead to drastic changes. Added optimisation, as in the case of [22], is a requirement in order to achieve the best possible forecast. This type of model can be used for a plethora of applications, where the underlying pattern is not easy to unearth.

*2) Recurrent Neural Networks (RNN):* An RNN is a modified version of the Feedforward Neural Network. It has an internal memory which can be used to process the sequence of the data. As the name suggests, an RNN is recurrent in nature, as the same function is performed for every single input. The output from an RNN depends on the nature of the input, and the previous computation. Because of the added internal memory, RNNs can be used for handwriting recognition, speech recognition and in other applications where the input is in the form of sequential data. Any RNN can be represented by Figure 2.

The approach provided by R. Madan and P.S. Mangipudi is quite beneficial for numerous purposes. Congestion control being one such area where forecasting may have a substantial impact [25]. This technology is lightweight and hence simple to use in areas such as data centres to enhance QoS parameters. Every model that was tested in [26] was in a position to take use of large time series datasets with extensive time series databases in relation. Seasonality may also be realised with the use of RNNs if no pre-seasonalization has occurred. When a dataset is able to take use of seasonal patterns that span over many series, all the accessible series may follow homogenous patterns and have identical durations in time. Leveraging them may provide a competitive advantage. In [27], a modified deep wavelet decomposition network was used to break down the input time series. In the article by Chen et al., a three-step technique was proposed, which included initialising, truncating, and message forwarding. To represent the interconnectedness of deconstructed raw time series data, as well as a technique to capture multiscale temporal connections, the results of this analysis were also shown in real time.

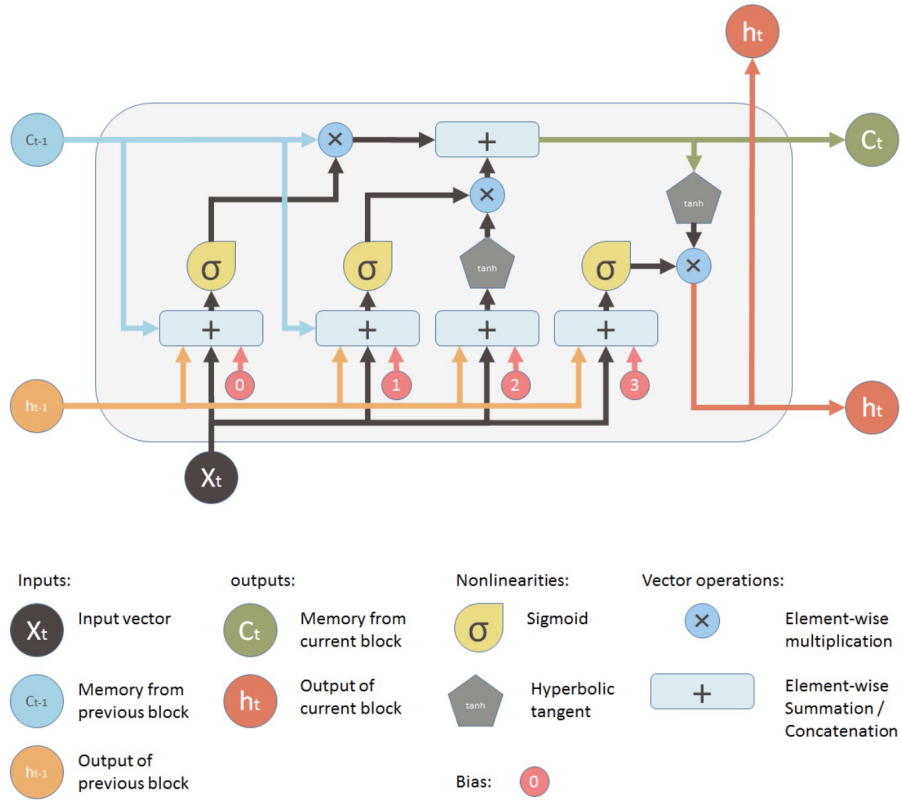

Fig. 3: A Long Short Term Memory Cell's Components [24]

*3) Long Short Term Memory:* RNNs have certain drawbacks, which are overcome with the help of LSTMs. RNNs are capable of only short term learning. They cannot retain very old data and tend to forget their earlier inputs. Every time new information is added, the current information is completely overwritten by RNNs. They cannot selectively decide which part of the information is important, and which is not. LSTMs are used to overcome all these problems. They are meant to learn long term dependencies, and have a higher prediction power. They can selectively add and remove information from their memory, depending on what is important and can even deal with the problem of exploding and vanishing gradients. There are four main parts to every LSTM cell, namely, the input gate, the output gate, the forget gate and the memory. The forget gate is used to decide what information is important and needs to be remembered, so that the rest can be forgotten. The input gate is used to add new information to the neuron cell, after it has been passed through some activation function. The output gate selects information that it deems as necessary, and then sends it as an output. A diagrammatic representation of the same can be seen in Figure 3.

In an effort to produce a forecasting model to target wind speed, researchers at K.U. Jaseena and Binsu C. Kovoor developed a decomposition-based methodology [29]. To arrive at the final prediction, a Bidirectional LSTM network was used to analyse frequencies, after which the forecasting outputs were procedure of cross validation cannot be carried out on time series data, as there is a temporal dependency and data flows in one direction. We cannot make multiple subsets of the data look at the future data and predict the past.

The method of cross validation that can be used for time series data is known as backtesting. In backtesting, an initial training period is specified, with a forecasting horizon. The model is trained and tested on those subsets. For the subsequent fold, the training set is expanded and the testing set is moved amalgamated to generate the final findings, putting this model very high on the set benchmark, based on precision and reliability. Nguyen et al. enlighten us on data-driven approaches for better and informed decision making as presented in [30]. On produced data, the LSTM based technique with the use of autoencoders performs better pertaining to detecting anomalies than the previously established LSTM model. With respect to sophisticated models, the LSTM-based forecasting model performs better than these models for various datasets. Hyper-parameters for hybridized algorithms that find irregularities are optimised further, with a suggestion provided on how to do so. In reference to a paper written by several authors in [31], they have suggested a process that is known as "Mid-Term Electric Load Forecasting." It utilises machine learning techniques like stochastic gradient descent to obtain a mapping as well as and representations of time series data. Seasonal associations can be secured with the help of LSTM owing to the inclusion of a better method of filtering with an emphasis on recollecting and retaining relevant data.

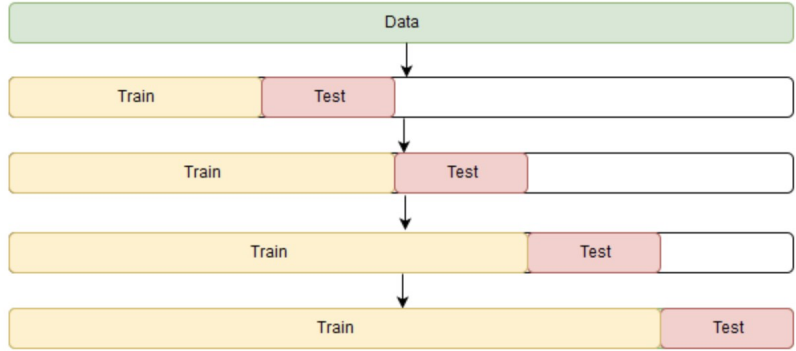

Fig. 4: Expanding Window Method for Backtesting [28]

IV. MODEL VALIDATION

The validation of forecasting models is an absolute necessity. This can be done with the help of various metrics. Most ML models face the issue of overfitting, and the error score will not represent the real learning of the model. The model would then perform excellently well on the training data, and can fail miserably when applied to the testing data. To overcome this issue, cross validation needs to be carried out, so that the true performance of the model can be brought to light. The standard forward. The training set can also be of a fixed size and can use the sliding window methodology. This process is repeated for multiple folds, and the final error score is the average of the error score for each fold. A diagrammatic representation of the expanding window method used in backtesting has been shown in Figure 4.

Various performance metrics can be used to calculate the error score so as to evaluate the model properly. With the help of these metrics, a quantitative sense of the performance of the model can be obtained and verified.

*1) Mean Absolute Error (MAE):* The Mean Absolute Error is an evaluation metric to measure the average magnitude of the errors, using the absolute value of the residuals. It is also called the Mean Absolute Deviation (MAD). The drawback of this metric is that it is scale dependent, and cannot be compared across datasets.

$$MAE = \frac{i}{n} \sum_{i=1}^{n} |e_i| \tag{7}$$

where, $n$ is the number of observations and $e_i$ is the error at the $i^{th}$ observation.

*2) Root Mean Squared Error (RMSE):* The Root Mean Squared Error is one of the most common error metrics in academics, that measures the average magnitude of the errors, using the standard deviation of the residuals. This metric is also scale dependent, gets biased because of large errors, and the final error score reflects the same.

$$RMSE = \sqrt{\frac{i}{n} \sum_{i=1}^{n} (e_i)^2} \qquad (8)$$

where, $n$ is the number of observations and $e_i$ is the error at the $i^{th}$ observation.

*3) Mean Absolute Percentage Error (MAPE):* This is one of the most commonly used error metrics in the business world. It is very similar to the MAE, but eliminates the scale dependency by normalising with respect to the true value. It calculates the error score in terms of the percentage of the error, when compared to the observed value. It is quite intuitive and easy to understand and interpret. The major drawback of this metric, is that it cannot be used when the actual values might have zeroes in them, as those zero values can never be made as denominators. Also, it can develop a bias towards overfitting and penalise it more as compared to underfitting.

$$MAPE = 100 * \frac{i}{n} \sum_{i=1}^{n} |\frac{e_i}{y_i}| \qquad (9)$$

where, $n$ is the number of observations and $e_i$ is the error at the $i^{th}$ observation, and $y_i$ is the true value of the data.

## V. CONCLUSION

Time Series Analysis can be used to see how a particular variable changes over time. Furthermore, it is often used to identify timely patterns and unearth trends that go unnoticed. Time Series Forecasting is the use of historical values and various patterns that are associated with that data, to predict future activity. The quality of the forecast depends on various factors, and can be considerably improved if the data has been preprocessed and prepared well. In this paper, we have provided a comprehensive description of the end to end process of TSF. Various methods of data preprocessing have been explained in depth. The various models used for forecasting have also been surveyed and presented in this paper. A summary of the same can be seen in Table I. With the help of this, a thorough understanding of the forecasting process can be obtained. Finally, we have also gone over various validation metrics and dived deep into backtesting and its working. Other error metrics and forecasting methods could have been considered in this paper, and that is something we will work on in the future, to improve the quality of the survey. In conclusion, this paper can be used to develop a good understanding of the overall process of TSA and TSF. It can also be used by anybody to get a grasp of the various state of the art models in the market today.

| Algorithm Name | Type of Model | Assumptions | Description | Forecast Type |
|---|---|---|---|---|
| SES | Statistical | No Trend, No Seasonality | Forecasts the future by taking an exponentially decreasing, weighted average of past occurrences, with the help of $\alpha$. | Linear |
| Holt Winters | Statistical | Trend & Seasonality are Both Present | Forecasts the future by taking weighted averages of the trend, seasonality and the level components, with the help of $\alpha$, $\beta$ and $\gamma$. | Seasonal |
| ARIMA | Statistical | Data is Stationary, Temporal Relationship in the Data | Forecasts the future by regressing the lagged values, differencing the data and taking a weighted mean of the historic errors. | Seasonal |
| Prophet | Statistical | Data can be Represented using the "Generalised Additive Model" | Forecasts the future by forecasting the trend and adding the appropriate seasonality, holidays and errors, i.e., applying a generalised additive model. | Seasonal |
| CNN | DL | Output is a Linear/Non-Linear Function of the Input | Convolves a kernel, with the same size as the width of the data, through the times series to find the relationship between the input and the output. | Seasonal |
| RNN | DL | Short Term Temporal Dependencies | Uses the same function recurrently, and passes on new information each time. Finds the short term relationships between the data. | Seasonal |
| LSTM | DL | Short Term & Long Term Spatio - Temporal Dependencies | Uses multiple functions to selectively choose a part of the input, selectively forget the current state and pick which part of the output to pass on. Finds even long term dependencies in the data. | Seasonal |

TABLE I: Summary of Various Algorithms for Time Series Forecasting

## References

[1] J. Brownlee, "What is time series forecasting?" Aug 2020. [Online]. Available: https://machinelearningmastery.com/time-series-forecasting

[2] A. Anish, "Time series analysis," Nov 2020. [Online]. Available: https://medium.com/swlh/time-series-analysis-7006ea1c3326

[3] [Online]. Available: https://madam-warlock.blogspot.com/2020/05/multiplicative-time-series-model.html

[4] C. C. Holt, "Forecasting seasonals and trends by exponentially weighted moving averages," *International journal of forecasting*, vol. 20, no. 1, pp. 5–10, 2004.

[5] E. Ostertagova and O. Ostertag, "Forecasting using simple exponential smoothing method," *Acta Electrotechnica et Informatica*, vol. 12, no. 3, p. 62, 2012.

[6] T. Booranawong and A. Booranawong, "Simple and double exponential smoothing methods with designed input data for forecasting a seasonal time series: In an application for lime prices in thailand." *Suranaree Journal of Science & Technology*, vol. 24, no. 3, 2017.

[7] W. Sulandari, Suhartono, Subanar, and P. C. Rodrigues, "Exponential smoothing on modeling and forecasting multiple seasonal time series: An overview," *Fluctuation and Noise Letters*, p. 2130003, 2021.

[8] C. Chatfield, "The holt-winters forecasting procedure," *Journal of the Royal Statistical Society: Series C (Applied Statistics)*, vol. 27, no. 3, pp. 264–279, 1978.

[9] L. Wu, X. Gao, Y. Xiao, S. Liu, and Y. Yang, "Using grey holt–winters model to predict the air quality index for cities in china," *Natural Hazards*, vol. 88, no. 2, pp. 1003–1012, 2017.

[10] W. Jiang, X. Wu, Y. Gong, W. Yu, and X. Zhong, "Holt–winters smoothing enhanced by fruit fly optimization algorithm to forecast monthly electricity consumption," *Energy*, vol. 193, p. 116779, 2020.

[11] M. V. De Assis, A. H. Hamamoto, T. Abrao, and M. L. Proença, "A game theoretical based system using holt-winters and genetic algorithm with fuzzy logic for dos/ddos mitigation on sdn networks," *IEEE Access*, vol. 5, pp. 9485–9496, 2017.

[12] G. E. Box, G. M. Jenkins, G. C. Reinsel, and G. M. Ljung, *Time series analysis: forecasting and control*. John Wiley & Sons, 2015.

[13] M. Ala'raj, M. Majdalawieh, and N. Nizamuddin, "Modeling and forecasting of covid-19 using a hybrid dynamic model based on seird with arima corrections," *Infectious Disease Modelling*, vol. 6, pp. 98–111, 2021.

[14] X. H. Nguyen *et al.*, "Combining statistical machine learning models with arima for water level forecasting: The case of the red river," *Advances in Water Resources*, vol. 142, p. 103656, 2020.

[15] G. Thejas, J. Soni, K. G. Boroojeni, S. Iyengar, K. Srivastava, P. Badrinath, N. Sunitha, N. Prabakar, and H. Upadhyay, "A multi-time-scale time series analysis for click fraud forecasting using binary labeled imbalanced dataset," in *2019 4th International Conference on Computational Systems and Information Technology for Sustainable Solution (CSITSS)*, vol. 4. IEEE, 2019, pp. 1–8.

[16] C. Olah, "Understanding lstm networks." [Online]. Available: https://colah.github.io/posts/2015-08-Understanding-LSTMs/

[17] S. J. Taylor and B. Letham, "Forecasting at scale," *The American Statistician*, vol. 72, no. 1, pp. 37–45, 2018.

[18] Y. Indulkar, "Time series analysis of cryptocurrencies using deep learning & fbprophet," in *2021 International Conference on Emerging Smart Computing and Informatics (ESCI)*. IEEE, 2021, pp. 306–311.

[19] G. Battineni, N. Chintalapudi, and F. Amenta, "Forecasting of covid19 epidemic size in four high hitting nations (usa, brazil, india and russia) by fb-prophet machine learning model," *Applied Computing and Informatics*, 2020.

[20] C. Guo, Q. Ge, H. Jiang, G. Yao, and Q. Hua, "Maximum power demand prediction using fbprophet with adaptive kalman filtering," *IEEE Access*, vol. 8, pp. 19236–19247, 2020.

[21] S. Mehtab, J. Sen, and S. Dasgupta, "Robust analysis of stock price time series using cnn and lstm-based deep learning models," in *2020 4th International Conference on Electronics, Communication and Aerospace Technology (ICECA)*. IEEE, 2020, pp. 1481–1486.

[22] N. Xue, I. Triguero, G. P. Figueredo, and D. Landa-Silva, "Evolving deep cnn-lstms for inventory time series prediction," in *2019 IEEE Congress on Evolutionary Computation (CEC)*. IEEE, 2019, pp. 1517–1524.

[23] I. E. Livieris, E. Pintelas, and P. Pintelas, "A cnn–lstm model for gold price time-series forecasting," *Neural computing and applications*, vol. 32, no. 23, pp. 17351–17360, 2020.

[24] S. Yan, "Understanding lstm and its diagrams," Nov 2017. [Online]. Available: https://blog.mlreview.com/understanding-lstm-and-its-diagrams-37e2f46f1714

[25] R. Madan and P. S. Mangipudi, "Predicting computer network traffic: a time series forecasting approach using dwt, arima and rnn," in *2018 Eleventh International Conference on Contemporary Computing (IC3)*. IEEE, 2018, pp. 1–5.

[26] H. Hewamalage, C. Bergmeir, and K. Bandara, "Recurrent neural networks for time series forecasting: Current status and future directions," *International Journal of Forecasting*, vol. 37, no. 1, pp. 388–427, 2021.

[27] D. Chen, L. Chen, Y. Zhang, B. Wen, and C. Yang, "A multiscale interactive recurrent network for time-series forecasting," *IEEE Transactions on Cybernetics*, 2021.

[28] M. S, "Using k-fold cross-validation for timeseries model selection," Mar 1960. [Online]. Available: https://stats.stackexchange.com/questions/14099/using-k-fold-cross-validation-for-time-series-model-selection

[29] K. Jaseena and B. C. Kovoor, "Decomposition-based hybrid wind speed forecasting model using deep bidirectional lstm networks," *Energy Conversion and Management*, vol. 234, p. 113944, 2021.

[30] H. Nguyen, K. P. Tran, S. Thomassey, and M. Hamad, "Forecasting and anomaly detection approaches using lstm and lstm autoencoder techniques with the applications in supply chain management," *International Journal of Information Management*, vol. 57, p. 102282, 2021.

[31] G. Dudek, P. Pełka, and S. Smyl, "A hybrid residual dilated lstm and exponential smoothing model for midterm electric load forecasting," *IEEE Transactions on Neural Networks and Learning Systems*, 2021.